\documentclass[lettersize,journal]{IEEEtran}
\newsavebox{\tempbox}

\usepackage{amsmath,amssymb,amsfonts}
\usepackage{algorithmic}
\usepackage{array}
\usepackage[caption=false,font=normalsize,labelfont=sf,textfont=sf]{subfig}
\usepackage{textcomp}
\usepackage{stfloats}
\usepackage{url}
\usepackage{verbatim}
\usepackage{graphicx}
\usepackage{cite}
\usepackage{booktabs, tabularx}
\usepackage{longtable}
\usepackage[]{hyperref}
\usepackage{xcolor}
\usepackage[normalem]{ulem}
\usepackage{balance}

\def\BibTeX{{\rm B\kern-.05em{\sc i\kern-.025em b}\kern-.08em
    T\kern-.1667em\lower.7ex\hbox{E}\kern-.125emX}}

\begin{document}
\title{Automated Construction of a Knowledge Graph of Nuclear Fusion Energy for Effective Elicitation and Retrieval of Information.}

\author{Andrea Loreti, Kesi Chen, Ruby George, Robert Firth, Adriano Agnello and Shinnosuke Tanaka.\\
\thanks{Andrea Loreti is with the UK Atomic Energy Authority, Culham Campus, Abingdon, OX14 3DB, UK and STFC Hartree Centre, Sci-Tech Daresbury, Keckwick, Daresbury, WA4 4AD.}
\thanks{Kesi Chen, Ruby George, Robert Firth and Adriano Agnello are with STFC Hartree Centre, Sci-Tech Daresbury, Keckwick, Daresbury, WA4 4AD.}
\thanks{Shinnosuke Tanaka is with the IBM Research, STFC Hartree Centre, Sci-Tech Daresbury, WA4 4AD.}
\thanks{This work was supported by the Hartree National Centre for Digital Innovation (STFC–IBM) and UKAEA–Hartree Centre collaboration.}}

\markboth{}%
{}
\maketitle

\begin{abstract}
In this document, we discuss a multi-step approach to automated construction of a knowledge graph, for structuring and representing domain-specific knowledge from large document corpora. We apply our method to build the first knowledge graph of nuclear fusion energy, a highly specialized field characterized by vast scope and heterogeneity. This is an ideal benchmark to test the key features of our pipeline, including automatic named entity recognition and entity resolution. We show how pre-trained large language models can be used to address these challenges and we evaluate their performance against Zipf’s law, which characterizes human natural language. Additionally, we develop a knowledge-graph retrieval-augmented generation system that uses multiple prompts with large language models to provide contextually relevant answers to natural-language queries, including complex multi-hop questions requiring reasoning across interconnected entities.
\end{abstract}


\section{Introduction}
\noindent Nuclear fusion has the potential to transform the global energy landscape by providing a sustainable source of carbon-free energy to our society. However, the achievement of nuclear fusion power presents numerous challenges that extend beyond the inherent complexities of nuclear physics and require multidisciplinary approaches, e.g., \cite{ITERtec,EuroFusion}. 

\noindent As the area of research continues to expand, the need for an efficient system of data management and knowledge retrieval has become increasingly critical \cite{Strand_2022,FAIR_2016,IMAS}. Addressing this challenge requires not only improving data accessibility but also enabling seamless interaction between human reasoning and machine processing.  

\noindent Retrieval-Augmented Generation (RAG) has recently emerged as a promising method to address such challenges, e.g., \cite{10.5555/3495724.3496517}. Standard RAG systems employ embeddings to perform semantic similarity search and retrieving relevant documents from text corpora. While effective for many tasks, embeddings often capture similarity rather than logical relevance and struggle with multi-hop reasoning, e.g., \cite{tang2024multihoprag}. These limitations can be mitigated by incorporating data models based on graph structures \cite{Brickley2004} or ontology frameworks \cite{McGuinness2004}. These can enhance the reasoning capabilities of RAG architectures by providing more structured representations of the underlying Knowledge Base (KB).

\noindent The development of a KB of fusion energy, will accelerate the development of more ''FAIR'' data sharing, disseminating and referencing \cite{Strand_2022, FAIR_2016}. Over time, such a system will foster collaborations between all partners in a fusion supply chain, helping to de-risk the design, development and construction of fusion power plants, while also reducing the barriers to understanding between field experts, funding agencies and policy makers.

\noindent In this work, we present a simple yet effective automated approach to generate a graph-based KB or Knowledge Graph (KG) of nuclear fusion energy starting from a large corpora of scientific documents and by leveraging the inference power of pre-trained Large Language Models (LLMs). Additionally, we developed a KG Retrieval-Augmented Generation (KG-RAG) machine. This combines the advanced linguistic capabilities of modern generative AI with domain-specific knowledge derived from previously unseen sources. Our study shows the potential of KG-RAG systems to enhance information elicitation and retrieval in domains characterized by complex and interconnected data sources.

\section{Related works}
\noindent The creation of a KB begins with the process of Information Extraction (IE) from a text corpus. To automate the learning of linguistic features, different trainable algorithms have been developed over the years. These include approaches based on statistical analysis \cite{osti_430781,Florian2004ASM}, Bayesian inference \cite{goldwater2007, Banko2008}, and machine learning algorithms, e.g., \cite{CRAVEN200069, carreras-marquez-2005-introduction}. Early methods often relied on precompiled heuristic rules and human supervision to extract linguistic patterns e.g., \cite{osti_430781} as well as on hand-annotated training sentences \cite{gildea-jurafsky-2002-automatic}. Other models employed ontologies to define classes and relationships of interest along with hand-labelled training texts, e.g., \cite{CRAVEN200069,ijcai2024p1039}. For these kinds of natural language processing tasks, manual efforts remain a significant bottleneck, especially when dealing with large text corpora.  

\noindent An essential task in IE is Named Entity Recognition (NER), which identifies entities specific to a particular domain within a given text. This process is often followed by Relation Extraction (RE), which establishes relationships between entities. Recurrent neural networks have been extensively used to accomplish these tasks, e.g., \cite{Wu2018} and \cite{Li2022}. More recently, pre-trained LLMs have demonstrated good IE capabilities across disciplines, in both zero- and few-shot scenarios \cite{agrawal2022large, Wei2023ZeroShotIE,hao-etal-2023-bertnet}. While their performance tends to decline when handling complex tasks \cite{Li2023EvaluatingCI} it can be recovered by breaking down high-level objectives into simpler, more manageable sub-tasks, each addressed through a targeted prompt \cite{Ashok2023, carta2023}. 

\noindent Over the years, various cross-disciplinary online KBs have been developed, e.g., Freebase or DBpedia \cite{Lehmann2015DBpediaA}. To the best of our knowledge, none of these KBs focuses on nuclear fusion energy, despite its societal relevance and the increasing interest shown by policymakers and the public. To address this gap, we embarked on the creation of the first KG for nuclear fusion energy.

\begin{figure*}[ht!]
    \centering
    \subfloat[]{\includegraphics[width=0.5\linewidth]{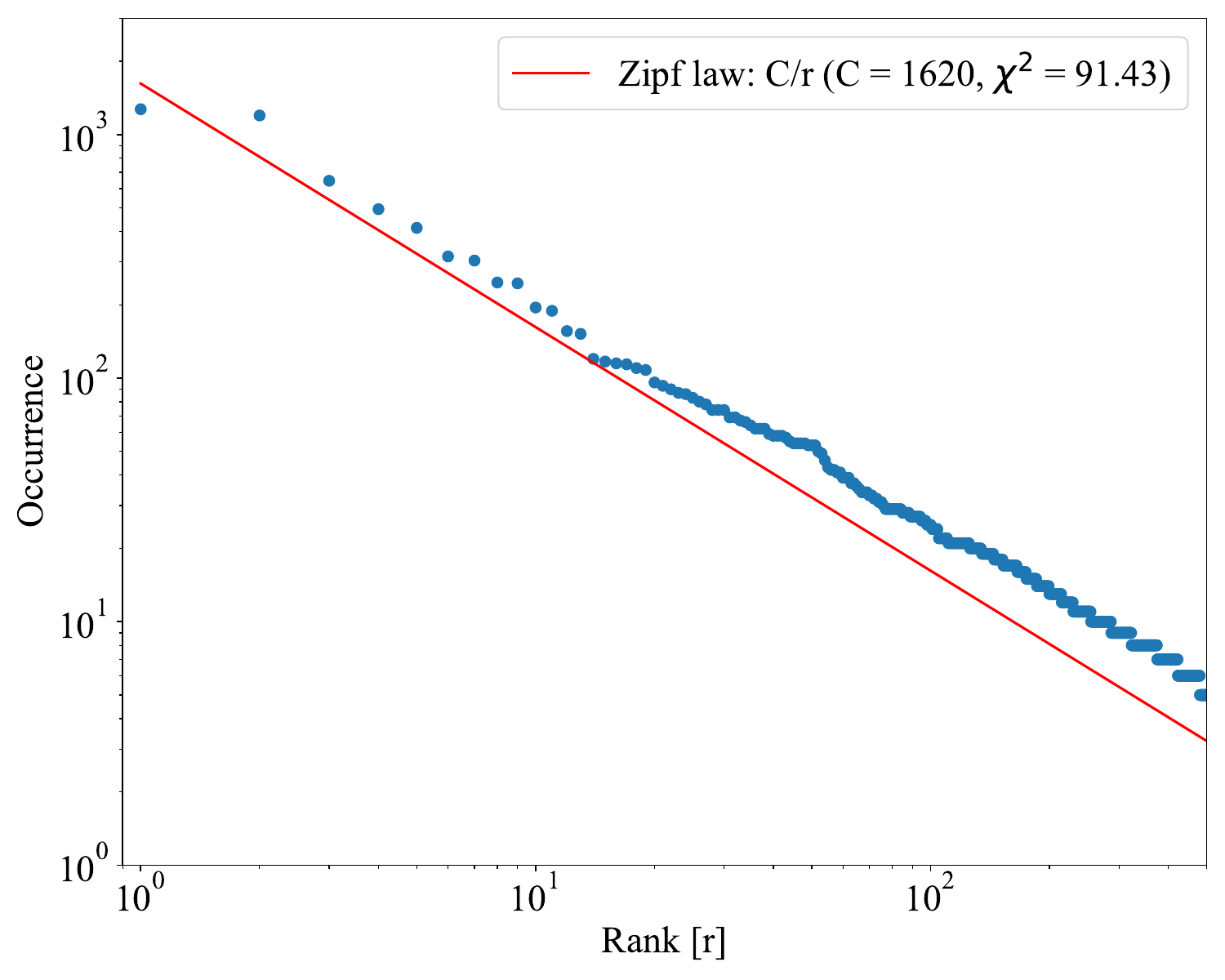}}
    \subfloat[]{\includegraphics[width=0.5\linewidth]{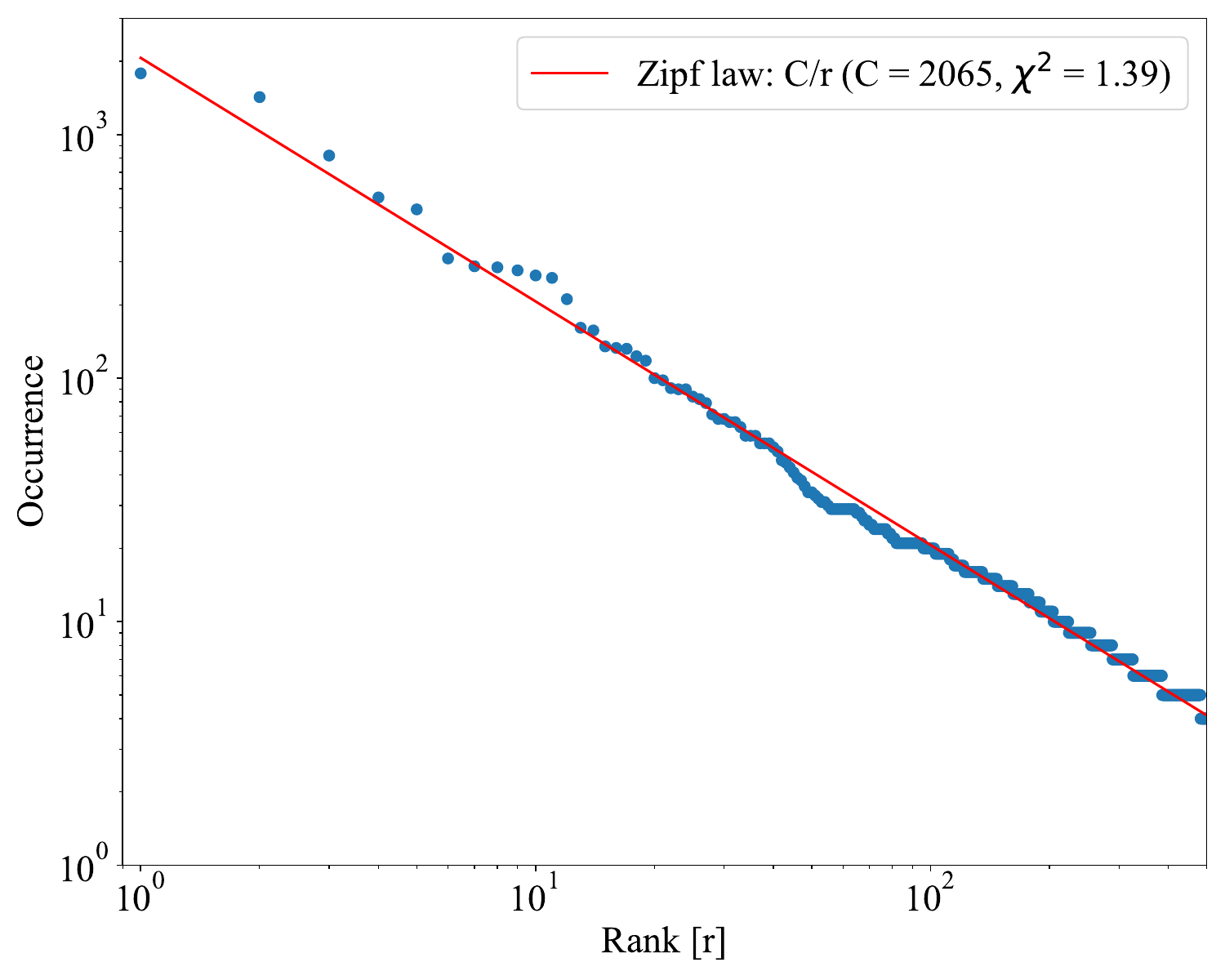}}
    \caption{Zipf's law applied to the top-ranked 500 single word entities in a case study of 349 abstracts: (a) before entity resolution, (b) after entity resolution process. The legend shows the Zipf parameter $(C)$ extracted from the fit and the normalized $\chi^2$ value.}
    \label{fig:Zipf}
\end{figure*}

\section{Methods}\label{section-method}
\noindent Our work can be divided in two main parts: the automated construction of a KG and the development of a KG-RAG. These are discussed in the next two sections.

\subsection{Automated construction of a KG}
\begin{table}[ht] 
\centering 
\caption{Data scoping} 
\label{DAQ-tab} 
\bigskip
\begingroup
\noindent
\begin{tabularx}{0.4\textwidth}{l@{\hspace{2\tabcolsep}}X}
\toprule&%
{%
  \begin{tabularx}{\hsize}[c]{X@{\hspace{2\tabcolsep}}c}%
  Key & Value%
  \setbox\tempbox\hbox{R-squared}%
  \xdef\Rsquzaredhsize{\the\wd\tempbox}%
  \end{tabularx}%
}%
\\
\midrule 
Summary: %
&%
{%
  \begin{tabularx}{\hsize}[c]{X@{\hspace{2\tabcolsep}}c}%
  Abstracts& 8358    \\
  Total words&  7799769     \\
  Time range& 1958-2024 \\
  \end{tabularx}%
}%
\\
\midrule 
Search patterns:%
&%
{%
  \begin{tabularx}{\hsize}[c]{X@{\hspace{2\tabcolsep}}c}%
  Tokamak & 4506\\
  Stellarator & 1632\\
  Inertial confinement &1202\\
  Nuclear fusion &328\\
  Fusion energy&355\\
  Deuterium-Tritium & 335\\
  \end{tabularx}%
}%
\\
\bottomrule
\end{tabularx}%
\endgroup
\end{table}

\begin{figure}[th!]
\centering
\includegraphics[trim=1cm 7.60cm 1cm 6.15cm, clip, scale=0.5]{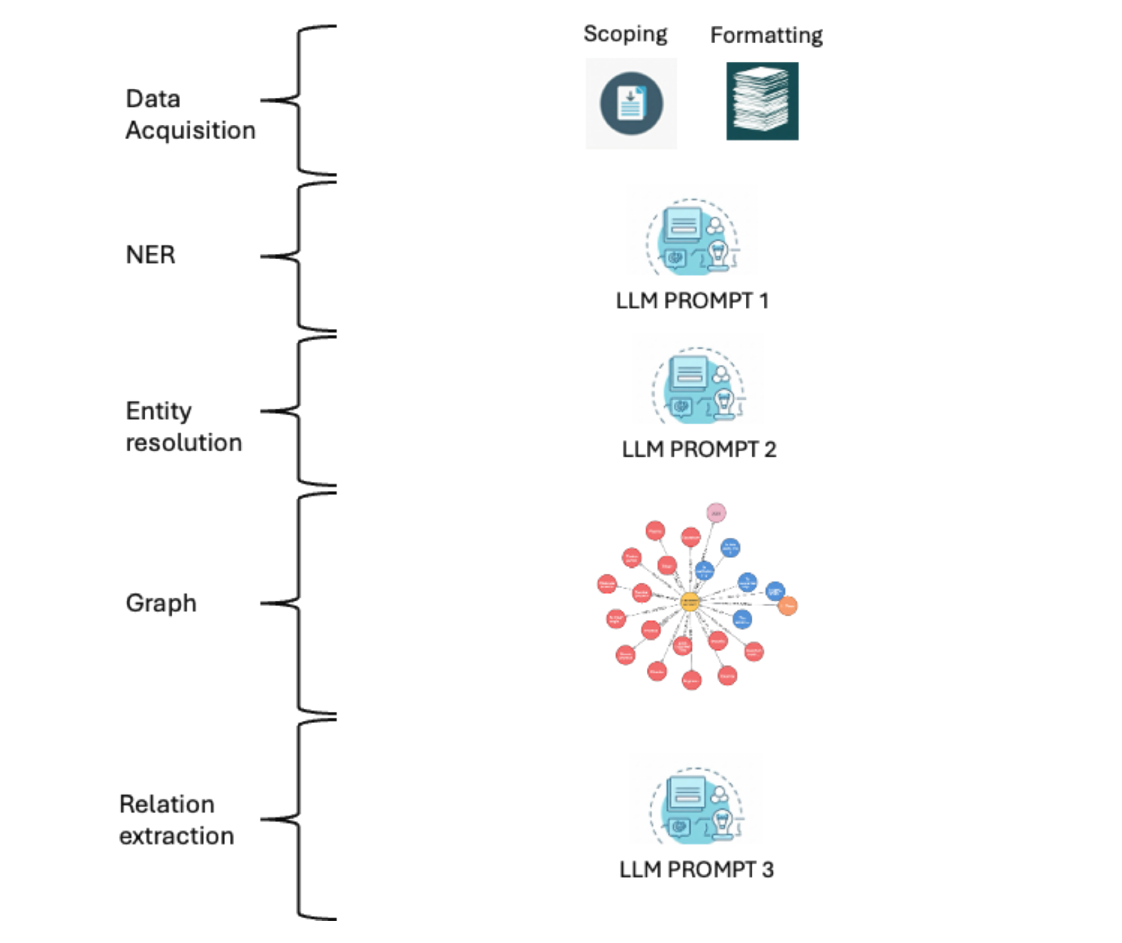}
\caption{Workflow for the automated construction of a KG, see text for more details.}
\label{pipeline_1}
\end{figure}

\begin{figure}[ht!]
\centering
\includegraphics[scale =0.2]{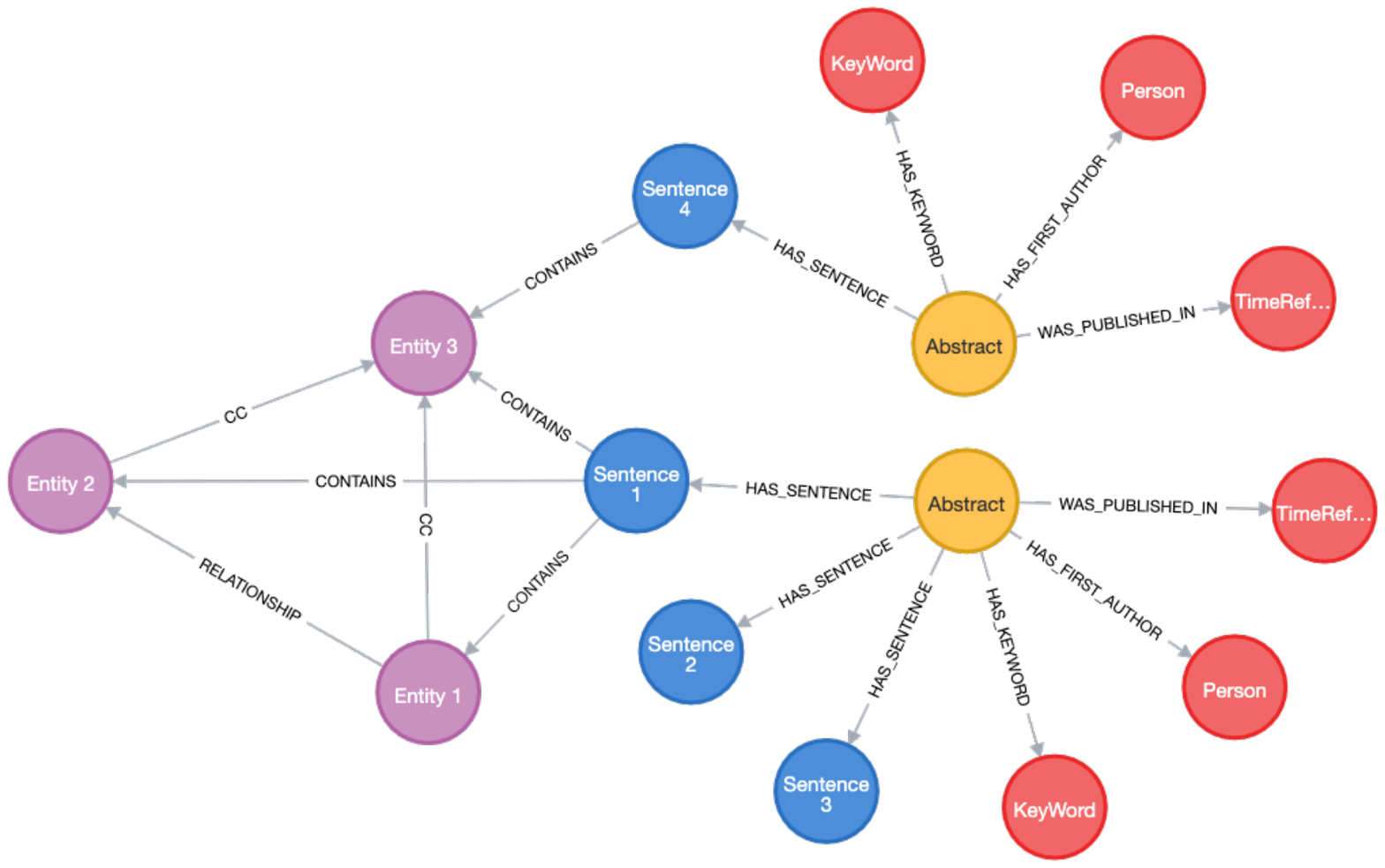}
    \caption{Example of KG accordingly to the graph architecture used in this work. In this diagram, (\textit{Entity 1, RELATIONSHIP, Entity 2}) form a triplet. Triplets have both [\text{CC}] and semantic relationships, see text for more details.}
    \label{fig:graph-example}
\end{figure}

\noindent In a graph, knowledge is stored in nodes linked together by edges, see for instance figure \ref{fig:graph-example}. Nodes encapsulate entities relevant to the domain while edges define relationships between them. 
\noindent For the automatic construction of our KG we used the pipeline shown in figure \ref{pipeline_1}. This is designed to be domain-agnostic and consists in the following: Data Acquisition (DAQ), NER, entity resolution, KG construction and RE.  

\paragraph{DAQ} this stage comprises data scoping and formatting. First, relevant data sources are identified by accessing the archive of scholarly publications at www.lens.org via dedicated API. The retrieved data are then structured in JSON format.

\noindent We downloaded 8358 scientific abstracts from journal articles related to nuclear fusion energy and associated metadata such as: authors, keywords, publication year and scholarly citation count. Abstracts were selected as they capture core terminology relevant for domain-level knowledge extraction. Articles were identified using title-based search patterns (Table \ref{DAQ-tab}). Irrelevant abstracts \footnote{The searching criteria included a small number of non-relevant papers, 8 out of 335 in the Deuterium–Tritium sample.} were filtered using an LLM-based validation step and a keyword filter.

\noindent Following the DAQ stage, the LLM performed NER on the selected texts.  

\paragraph{NER} can be accomplished in few- or zero-shot mode depending on whether examples of entities are passed to the prompt or not, \cite{agrawal2022large, hao-etal-2023-bertnet, Ashok2023, carta2023}. In the present work, we proposed a set of category types that are as exclusive as possible while covering a large portion of relevant entities (see Appendix). While this introduces a degree of domain specificity it also enables standardized, cross-model categorization that may be used during the retrieval stage. Classification accuracy is estimated through human assessment of LLM-assigned categories on randomly selected entities. This estimate depends on the specific choice of category types. It also represents an upper bound on accuracy, since an entity may be validly assigned to multiple categories. \noindent  We used Llama3.1 or Llama 3.3 by Meta, deployed on Amazon Web Services (AWS) equipped with 405 and 70 billion parameters, respectively \footnote{meta.llama3-1-405b-instruct-v1:0 and meta.llama3-3-70b-instruct-v1:0}. These models correctly classified $\lesssim$90\% of the entities. The accuracy was seen to decrease to $\lesssim70$\% for smaller LLMs such as Llama3.1 8B. Notably, the landscape of LLMs has evolved rapidly since the beginning of this study, with the emergence of both compact dense models and sparse mixture‑of‑experts architectures that achieve comparable text‑processing performance to much larger dense models at significantly reduced computational costs, e.g., \textsc{Phi‑3}, \textsc{Gemma~2}, and \textsc{DeepSeek‑V2/V3}.

\paragraph{Entity resolution} the main sources of error in NER are entity hallucinations (LLM-inferred entities not present in the text), missed entities, and misclassifications. To reduce these errors, we narrowed the LLM’s focus by splitting text into individual sentences. We then applied entity resolution to standardize entity forms (converting to singular or lowercase where appropriate\footnote{The spaCy software package was used for this task}), exclude numeric-only entities, and remove hallucinations.
\noindent As part of entity resolution, we performed an additional LLM pass to expand acronyms and standardize chemical element names (this step may require user supervision). Entity resolution improves consistency and reduces redundancy in the final KG. \footnote{For instance, it allowed to reduce the number of nodes from 4330 to 3632, and interlinks from 31076 to 25004 in a KG constructed from 334 abstracts.}

\noindent The effect of entity resolution can be appreciated by looking at the plots in figure \ref{fig:Zipf}. These show the frequency of the 500 top-ranked single word entities fitted by the Zipf's law, before and after de-noising stage (entity resolution). 

\noindent The Zipf's law is an empirical principle observed in linguistics. It states that in a given text corpus, the frequency $(f)$ of any word is inversely proportional to its rank $(r)$ in the frequency table: $f = C/r$. Applying Zipf’s law provides a quantitative benchmark to assess the naturalness and linguistic plausibility of the extracted entities. Deviations from its trend may reveal noise, or redundancy in the NER. After entity resolution, the frequency distribution of our extracted entities aligns better with Zipf’s law.

\paragraph{KG construction} for this step we have used the Neo4j framework. The graph structure is summarized in table \ref{Nodes-Relationship-table} and includes nodes for abstracts, first authors, publication years, and key words, as well as individual sentences and entities. These nodes are interconnected through a set of predefined relationships, also summarized in table \ref{Nodes-Relationship-table} and illustrated in figure \ref{fig:graph-example}:

\begin{table}[h!] 
\centering 
\caption{Nodes and relationships properties} 
\label{Nodes-Relationship-table} 
\bigskip
\begingroup
\noindent
\begin{tabularx}{\columnwidth}{l X X}  
\toprule
\textbf{Label} & \textbf{Key} & \textbf{Value} \\  
\midrule 
Abstract  
    & \textbf{name}          & Abstract title   \\
    & \textbf{text}          & Abstract text    \\
    & \textbf{url}           & URL              \\
    & \textbf{citationCount} & Number of citations \\
\midrule 
Sentence
    & \textbf{name}       & Sentence text \\
    & \textbf{embeddings} & Vector embedding \\
\midrule 
Entity/Person/\\
TimeReference/KeyWord  
    & \textbf{name}       & Entity name \\
    & \textbf{types} & Entity category type \\
    & \textbf{edges} & Number of edges \\
\bottomrule
\textbf{Label} & \textbf{Start/end} & \textbf{Value} \\  
\midrule 
HAS\_FIRST\_AUTHOR
    & \textbf{Abstract/Person}          & \\
WAS\_PUBLISHED\_IN 
    & \textbf{Abstract/TimeReference}          & \\
HAS\_KEYWORD
    & \textbf{Abstract/KeyWord}          &  \\

HAS\_SENTENCE
    & \textbf{Abstract/Sentence}          &  \\
CONTAINS
& \textbf{Sentence/Entity}          &  \\
\midrule
CC
& \textbf{Entity/Entity}          & weight \\
& & text (optional)\\
\bottomrule
\end{tabularx}%
\endgroup
\end{table}

\noindent The link $[\text{CC}]$ establishes the co-occurrence of two entities in the same sentence by assigning a weight to this relationship given by the number of times this co-occurrence happens.

\paragraph{Relation extraction} beyond the $[\text{CC}]$ link, semantic relationships, such as predicative or causal connections, are inferred via a LLM pass. These are introduced only for highly recurrent nodes identified by filtering entities outside a three-standard-error range in bond strength and edge count.\footnote{Sentences containing both candidate nodes are ranked by cosine similarity, and the top six are used by the LLM to infer relationships.}

\noindent A qualitative inspection of the LLM‑inferred relationships reveals two main limitations. 
First, filtering entities by recurrence does not always guarantee semantically meaningful triplets, as highly frequent entities may correspond to generic terms that do not require explicit relational grounding (e.g., \{\textit{optimization}, \textit{applied\_to}, \textit{stellarator}\}). 
Second, extracted relations often take the form of full sentences rather than concise predicative or causal links, despite explicit prompting.  While this behaviour does not imply factual inaccuracies, it introduces redundancy and reduces compactness. 
We suggest that these effects could be mitigated through stricter prompt constraints or post‑processing filters.

\subsection{KG-RAG}
\begin{figure}[ht!]
\includegraphics[trim=0.6cm 5.75cm 0.5cm 6.5cm, clip, scale=0.45]{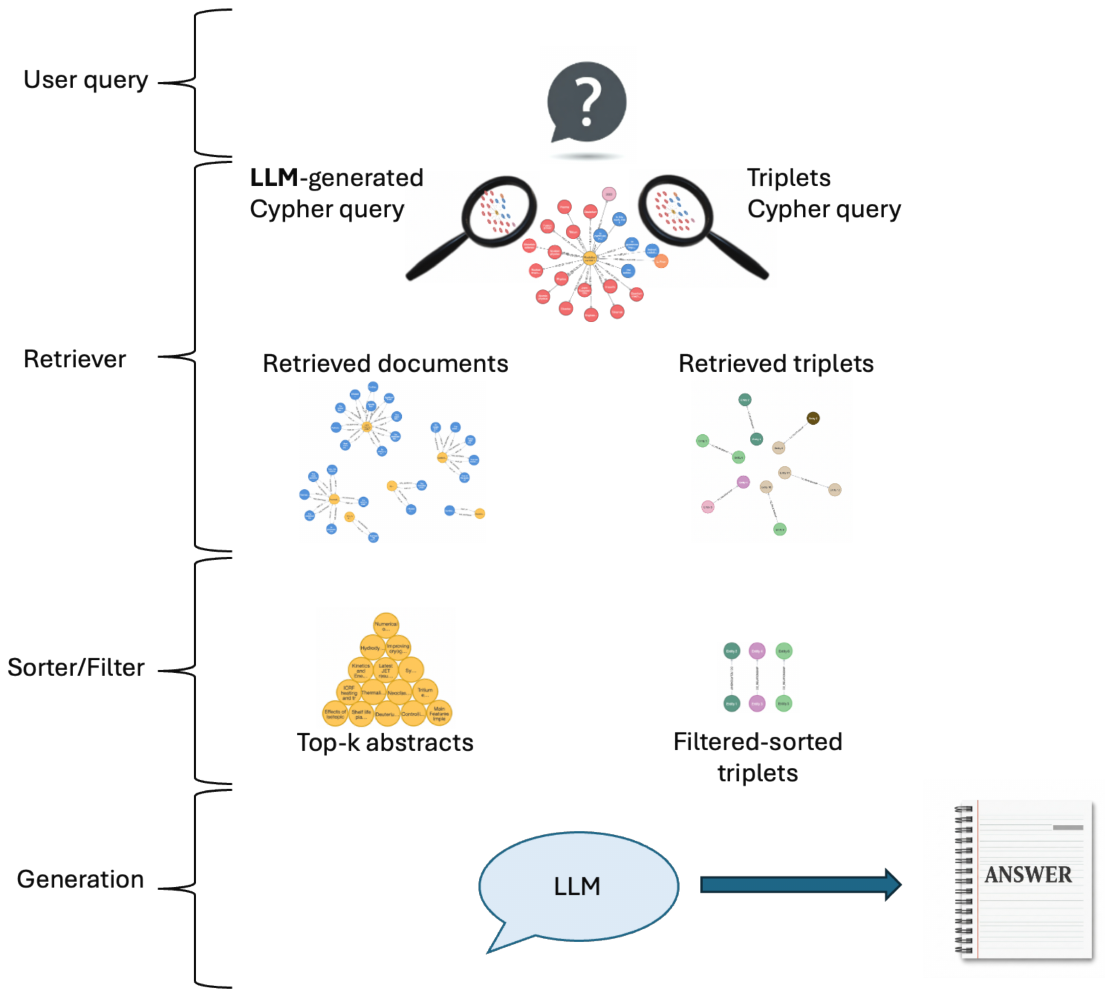}
\caption{Workflow of the KG-RAG, see text for more details.}
\label{pipeline_2}
\end{figure}

\begin{figure}[]
    \centering
    \includegraphics[trim=0cm 1.5cm 0cm 1.5cm, clip,width=1\linewidth]{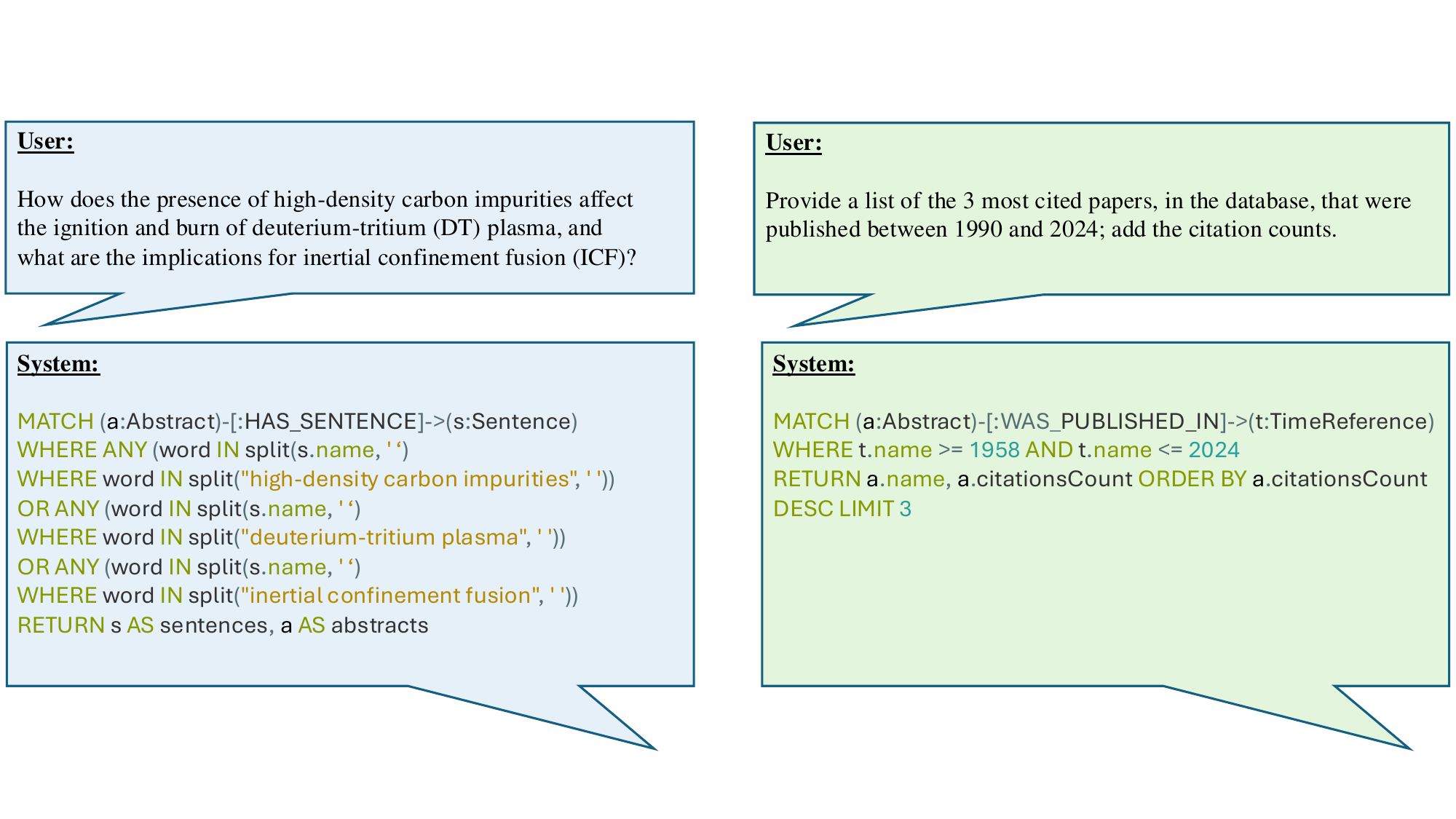}
    \caption{Two examples of LLM-generated Cypher queries, single-hop (left) and multi-hop (right), derived from user queries expressed in natural language. For generating Cypher queries, LLM was provided via prompt with the graph structure in table \ref{Nodes-Relationship-table}.}
    \label{fig:Cypher-queries}
\end{figure}

\noindent The automated search of the KG for retrieval and generation purposes is accomplished by using the pipeline in figure \ref{pipeline_2}. In the first step, the user submits a question and specifies the number \textit{k} of top documents to retrieve. A first LLM pass extracts relevant entities and embeds them into a Cypher query. This is a language designed for working with graph databases. Figure \ref{fig:Cypher-queries} shows two LLM-generated Cypher queries generated from user inputs. We consider both single-hop and multi-hop queries that can be answered by using information from a single node or multiple interconnected nodes, respectively.

\noindent We perform full-text search to find sentences matching entities from the user's query. These are ranked by semantic similarity using a \textit{sorter} module, which computes cosine similarity between sentences and query embeddings \footnote{Embeddings are generated with the model https://huggingface.co/sentence-transformers/all-mpnet-base-v2, that maps text into a 768-dimensional space.}. The top-k abstracts containing the most relevant sentences are then used to prompt the LLM for answer generation.

\noindent The retriever also searches the graph for triplets (\textit{subject–relationship–object}). These are filtered to include only those containing entities from the user’s query or verbs with matching lemmas. If no match is found, triplets are ranked by subject–object frequency. The selected triplets are then passed to the LLM for answer generation.

\section{Results}
\noindent In this study, we accomplished the automated construction of a KG of nuclear fusion energy that comprehensively represents the domain knowledge contained in 8358 scientific abstracts. To the best of our knowledge, this is the first KG in the domain of nuclear fusion energy. The graph structure is summarized in tables \ref{Nodes-Relationship-table} and contains 108811 nodes and 718335 links (for the dataset of categorized entities see \cite{FusionKnowledgeDataset}).

\noindent Assessing the performance of a RAG (or KG-RAG) system requires evaluating both retrieval accuracy and the faithfulness and correctness of generated responses, e.g., \cite{saad-falcon-etal-2024-ares} and \cite{Yu_2025}. An exhaustive study of RAG performances is beyond the scope of this work. However, we conducted two evaluation experiments. First, we randomly selected $10$ abstracts and used ChatGPT to generate one question each, simulating queries from field experts. We then checked whether the right abstract appeared among the retrieved sources. In this experiment, the correct abstract was consistently ranked as the top result. 

\noindent Next, we generated a set of $50$ questions using a system prompt designed to simulate queries from different kinds of personas \footnote{We used Ragas:  https://docs.ragas.io/en/v0.1.21/index.html}. These questions were less specific, shorter, and contained typos or colloquial phrases. The retrieval accuracy for this broader set was $\simeq 50\%$. However, in all but one case, the retrieved documents remained highly relevant to the corresponding queries showing the system capabilities to consistently surface relevant information.

\noindent It must be noted that for all generated answers, a list of correct references is provided which links to the original text sources via URL. This result addresses a well-known problem associated with LLMs, specifically the issue of referencing the text source used in answer generation. In fact, even top-performing models currently available have been found to provide unsupported statements or incorrect, conflicting citations \cite{wu2024llmsciterelevantmedical}. 

\noindent For what is concerned with the generative process, this primarily consists in summarizing the retrieved top-ranked documents and does not add extra information from knowledge acquired during the pre-training, unless explicitly prompted to do so, see example 1. If, on the one hand, these guardrails may limit the creative scope of generative AI, they also help to prevent LLMs from hallucinating and producing fictitious contextual answers, thus addressing a well-known issue affecting these models, e.g., \cite{10.1145/3571730}.
\noindent Although accurate, answers generated by summarizing retrieved relevant documents may lack abstraction. Tasks such as summarizing the entire knowledge graph or extracting time-specific information often require multi-hop reasoning that goes beyond the content of individual retrieved documents. In a simple RAG system, achieving this level of abstraction can be difficult, as retrieval is limited to the top-\textit{k} documents selected via embedding-based similarity. In a KG-RAG framework, this abstraction can be achieved by explicitly leveraging graph relationships. For instance, by using the links in (Table \ref{Nodes-Relationship-table}) and LLM-generated semantic relations, KG-RAG gains an initial abstraction layer, as shown in examples 2 and 3.

\begin{itemize}
\item \textbf{Example 1}\\
\textbf{User}: \textit{"How does the presence of high-density carbon impurities affect the ignition and burn of deuterium-tritium (DT) plasma, and what are the implications for inertial confinement fusion (ICF)?"}\\ 
\textbf{System}: "The presence of high-density carbon impurities has a negative effect on the ignition and burn of deuterium-tritium (DT) plasma in inertial confinement fusion (ICF) targets..."\\
\textbf{Sources}: \cite{10.1063/1.5087298}, \cite{10.1063/5.0008231} , \cite{10.1063/1.1311940}.
\item \textbf{Example 2} \\ 
\textbf{User}: \textit{"Provide a list of the 3 most cited papers, in the database, published between 1958 and 2024; add the citation counts."}\\
\textbf{System}: 
Based on the text, here is a list of the top 3 most cited papers along with their citation counts:
\begin{enumerate}
\item "Development of the indirect‐drive approach to inertial confinement fusion and the target physics basis for ignition and gain"	2267, \cite{Lindl_1995}
\item "Neoclassical transport of impurities in tokamak plasmas"	1077, \cite{Hirshman_1981}
\item "Measurements of microturbulence in tokamaks and comparisons with theories of turbulence and anomalous transport"	768, \cite{Liewer_1985}
\end{enumerate}
\item \textbf{Example 3}\\
\textbf{User}: \textit{"Can you summarize the content of the database?"}\\
\textbf{System}:
The data set appears to be a collection of sentences related to nuclear physics, specifically focusing on topics such as fusion reactions, plasma, and neutron interactions. The sentences describe various relationships between different entities, including deuterium, tritium, neutrons, and other particles, as well as concepts like ignition, implosion, and radiation.
The data set seems to be a compilation of information from various sources, possibly research papers or scientific articles, and covers a range of topics, including...
\end{itemize}

\section{Conclusion}
In this work, we developed a modular, generalizable, and domain-agnostic multi-step approach to automated construction of a KG. To demonstrate feasibility of our approach, we applied it to the nuclear fusion domain, where minimal specific adaptations (like acronym expansion, and category selection) were required. 
\noindent In our pipeline, we leveraged the inference power of LLMs to extract entities and relationships from 8358 scientific abstracts in the field, creating a KG that effectively represents its domain knowledge. This comprehensive representation is not only beneficial for data storage but also serves as a foundation for the development of a KG-RAG system. Our results highlight the system’s abstraction capabilities over a complex domain knowledge, demonstrating its potential for enhancing information retrieval and knowledge representation in specialized fields.

\section{Acknowledgments}
The authors used generative AI to improve the clarity of the manuscript. All ideas, logical structure, methodology, and conclusions are the result of the authors' own work.

\appendix
\section{Appendix A: list of category types}\label{sec-Appendix-A}
List of the 28 category types used in our analysis and during NER: Concept, Nuclear Fusion Experimental Facility, Nuclear Fusion Technique, Nuclear Fusion Device Type, Nuclear Fusion System Component, Nuclear Fusion System Configuration, Experimental Apparatus, Physical Process, Physics Entity, Field Configuration, Particle and Subatomic particles, Chemical Element or Compound, Plasma Property, Plasma Event, Plasma Region, Plasma Dynamics and Behavior, Detection and Monitoring Systems, Control Systems,  Theory and Calculation, Software and Simulation, Time Reference, Country and Location, Facility or Institution, Person, Safety Feature and Regulatory Standard, Database, Scientific Publication and Citation, Research Field.

\bibliographystyle{ieeetr}
\bibliography{Biblio}

\end{document}